\documentclass{article} 
\pdfoutput=1
\usepackage{nips13submit_e,times}
\usepackage{hyperref}
\usepackage{url}
\usepackage{amssymb}
\usepackage{graphicx}
\nipsfinalcopy
\title{cuDNN: Efficient Primitives for Deep Learning}

\author{
Sharan Chetlur, Cliff Woolley, Philippe Vandermersch, Jonathan Cohen, John Tran \\
NVIDIA\\
Santa Clara, CA 95050 \\
\texttt{\{schetlur, jwoolley, philippev, jocohen, johntran\}@nvidia.com} \\
\And
Bryan Catanzaro \\
Baidu Research\\
Sunnyvale, CA 94089 \\
\texttt{bcatanzaro@baidu.com} \\
\And
Evan Shelhamer \\
UC Berkeley \\
Berkeley, CA 94720 \\
\texttt{shelhamer@eecs.berkeley.edu}
}

\begin{document}

\maketitle

\begin{abstract}
We present a library of efficient implementations of deep learning primitives.
Deep learning workloads are computationally intensive, and optimizing their kernels is difficult and time-consuming.
As parallel architectures evolve, kernels must be reoptimized, which makes maintaining codebases difficult over time.
Similar issues have long been addressed in the HPC community by libraries such as the Basic Linear Algebra Subroutines (BLAS) \cite{NetlibBLAS}.
However, there is no analogous library for deep learning.
Without such a library, researchers implementing deep learning workloads on parallel processors must create and optimize their own implementations of the main computational kernels, and this work must be repeated as new parallel processors emerge.
To address this problem, we have created a library similar in intent to BLAS, with optimized routines for deep learning workloads.
Our implementation contains routines for GPUs, although similarly to the BLAS library, these routines could be implemented for other platforms.
The library is easy to integrate into existing frameworks, and provides optimized performance and memory usage.
For example, integrating cuDNN into Caffe, a popular framework for convolutional networks, improves performance by 36\% on a standard model while also reducing memory consumption.
\end{abstract}

\section{Introduction}
Deep neural networks have been successful at solving many kinds of tasks \cite{Bengio2012}.
Parallel processors such as GPUs have played a significant role in the practical implementation of deep neural networks.
The computations that arise when training and using deep neural networks lend themselves naturally to efficient parallel implementations.
The efficiency provided by these implementations allows researchers to explore significantly higher capacity networks, training them on larger datasets \cite{Coates2013}.
This has led to greatly improved accuracy on tasks such as speech recognition and image classification, among others.
For example, the majority of entries in the 2014 ILSVRC challenge \cite{ILSVRC2014} use GPUs to implement deep neural networks, as pioneered by \cite{Krizhevsky2012}, and improved by many others, including \cite{Sermanet2013} \cite{Szegedy2014}.
Deep neural networks for speech recognition have also benefited from parallel implementations on GPUs \cite{Hinton2012} \cite{Dahl2012} \cite{Maas2014}.

Convolutional Neural Networks (CNNs) \cite{Lecun1998} are an important and successful class of deep networks.
Convolutional neural networks are computed using dense kernels that differ from traditional dense linear algebra routines.
Accordingly, modern deep learning frameworks, such as Torch7 \cite{Collobert2011}, Theano \cite{Bergstra2010}, and Caffe \cite{Jia2014} feature suites of custom kernels that implement basic operations such as tensor convolutions, activation functions and pooling.
These routines represent the bulk of the computation when training a CNN, and thus account for the majority of its execution time.
The deep learning community has been successful in finding optimized implementations of these kernels, but as the underlying architectures evolve, these kernels must be re-optimized, which is a significant investment.
Optimizing these kernels requires a deep understanding of the underlying processor architecture, with careful scheduling of data movement, on-chip memory placement, register blocking, and other optimizations in order to get acceptable performance.

We believe that providing a library of optimized routines for these computations will provide several important benefits.
Firstly, deep learning frameworks can focus on higher-level issues rather than close optimization of parallel kernels to specific hardware platforms.
Secondly, as parallel architectures evolve, library providers can provide performance portability, in much the same way as the BLAS routines provide performance portability to diverse applications on diverse hardware.
Thirdly, a clearer separation of concerns allows specialization: library providers can take advantage of their deep understanding of parallel architectures to provide optimal efficiency.
Our goal is to make it much easier for deep learning frameworks to take advantage of parallel hardware.

Our library addresses this goal by providing a flexible, easy-to-use C-language API for deep learning workloads that integrates neatly into existing frameworks.
It can provide immediate efficiency gains, and it is rigorously tested and maintained in order to be reliable and performant across a range of different processor architectures.
Importantly, our library is designed to use the minimum possible amount of auxiliary memory, which frees up scarce memory for larger models and datasets.
We also optimize performance across a wide range of potential use cases, including small mini-batch sizes.

\section{Library}
One of the primary goals of cuDNN is to enable the community of neural network frameworks to benefit equally from its APIs.
Accordingly, users of cuDNN are not required to adopt any particular software framework, or even data layout.

Rather than providing a layer abstraction, we provide lower-level computational primitives, in order to simplify integration with existing deep learning frameworks, each with their own abstractions.

The bulk of the API is dedicated to functions that perform primitive operations on data stored in user-controlled buffers.
By keeping the API low-level, the library integrates simply into other frameworks.

cuDNN supports forward and backward propagation variants of all its routines in single and double precision floating-point arithmetic.
These include convolution, pooling and activation functions.
The library allows variable data layout and strides, as well as indexing of sub-sections of input images.
It also includes a set of auxiliary tensor transformation routines that allow for the easy manipulation of 4d-tensors.

\subsection{Overview and Handles}
The library exposes a host-callable C language API, but requires that input and output data be resident on the GPU, analogously to cuBLAS.
The library is thread-safe and its routines can be called from different host threads.
Convolutional routines for the forward and backward passes use a common descriptor that encapsulates the attributes of the layer.
Tensors and Filters are represented in opaque descriptors, with the flexibility to specify the tensor layout using arbitrary strides along each dimension for tensors.

\subsubsection{Spatial Convolutions}
The most important computational primitive in convolutional neural networks is a special form of batched convolution.
The parameters governing this convolution are listed in table \ref{tab:conv_param}. In this section, we describe the forward form of this convolution - the other forms necessary for backpropagation are closely related.

\begin{table}
\centering
\begin{tabular}{ll}
Parameter & Meaning \\ \hline
$N$ & Number of images in mini-batch \\
$C$ & Number of input feature maps \\
$H$ & Height of input image \\
$W$ & Width of input image \\
$K$ & Number of output feature maps \\
$R$ & Height of filter kernel \\
$S$ & Width of filter kernel \\ \hline
$u$ & Vertical stride \\
$v$ & Horizontal stride \\
$pad\_h$ & Height of zero-padding \\
$pad\_w$ & Width of zero-padding \\
\end{tabular}
\caption{Convolutional parameters}
\label{tab:conv_param}
\end{table}

There are two inputs to the convolution: $D \in \mathbb{R}^{NCHW}$, which forms the input data, and $F \in \mathbb{R}^{KCRS}$, which forms the convolutional filters.
The input data ranges over $N$ images in a mini-batch, $C$ input feature maps, $H$ rows per image, and $W$ columns per image.
The filters range over $K$ output feature maps, $C$ input feature maps, $R$ rows per filter, and $S$ columns per filter.
The output is also a four-dimensional tensor $O \in \mathbb{R}^{NKPQ}$, where $N$ and $K$ were as defined previously, and $P=f(H, R, u, pad\_h)$, $Q=f(W, S, v, pad\_w)$, meaning that the height and width of the output images depends on the image and filter height and width, along with padding and striding choices.
The striding parameters $u$ and $v$ allow the user to reduce the computational load by computing only a subset of the output pixels.
The padding parameters allow users to specify how many rows or columns of $0$ entries are appended to each image.
MATLAB's ``valid'' convolution mode corresponds to setting $pad\_h=0, pad\_w=0$, while MATLAB's ``same'' convolution mode corresponds to $pad\_h=\left\lfloor\frac{R}{2}\right\rfloor, pad\_w=\left\lfloor\frac{S}{2}\right\rfloor$,
and MATLAB's ``full'' convolution mode corresponds to $pad\_h=R-1, pad\_w=S-1$.

More specifically,
\begin{equation}
f(H, R, u, pad\_h) = \left\lceil\frac{H - R + 1 + 2pad\_h}{u}\right\rceil
\end{equation}

Define an accessing function to account for striding, padding, and inverting for the convolution:
\begin{equation}
g(p, u, R, r, pad\_h) = p\cdot u + R - r - 1 - pad\_h
\end{equation}

Then, the forward convolution evaluates $O[n, k, p, q] \; \forall n \in [0, N), \forall k \in [0, K), \forall p \in [0, P), \forall q \in[0, Q)$.
For convenience, define $D_0$ as a zero-extended version of $D$.
\begin{equation}
O[n, k, p, q] = \sum_{c=0}^{C-1} \sum_{r=0}^{R-1} \sum_{s=0}^{S-1}
    F[k, c, r, s] \cdot D_0[n, c, g(p, u, R, r, pad\_h), g(q, v, S, s, pad\_w)]
\label{eq:conv}
\end{equation}

As can be seen from Equation \ref{eq:conv}, computing the convolution involves a seven-way nested loop, with four independent loops and three accumulation loops.
There are many ways of implementing this computation, some of which we will discuss in the next section.

cuDNN's convolutional routines incorporate implementations of both the convolution as well as the cross-correlation variants of these functions.
  These functions support user-defined strides along each dimension of the input and output tensors.
  This is important because different frameworks store tensors using different memory layouts; for example, some frameworks interleave feature maps, while others keep them separate.
  cuDNN allows the user to specify memory layout, which makes it much simpler to integrate into existing frameworks.
  cuDNN's routines also have a mode to either return the raw gradients or to accumulate them in a buffer as needed for models with shared parameters or a directed acyclic graph structure.

\subsection{Other functions}
cuDNN also provides other commonly used functions for deep learning. For example, it provides three commonly used neuron activation functions; Sigmoid, Rectified Linear and Hyperbolic Tangent.
  It provides a softmax routine, which by default uses the numerically stable approach of scaling each element to avoid overflow in intermediate results.
  Softmax may be computed per image across the feature map, height and width dimensions or per spatial location, per image across the feature map dimension.
  cuDNN provides average and max pooling operations, as well as a set of tensor transformation routines, such as those that add tensors, with optional broadcasting.
  The goal of providing these functions is to reduce the amount of parallel code that is required for deep learning frameworks, by providing flexible, well-optimized versions of these commonly used functions.
  With cuDNN, it is possible to write programs that train standard convolutional neural networks without writing any parallel code, but simply using cuDNN and cuBLAS.

\section{Implementation}
The majority of functions that cuDNN provides have straightforward implementations.
The convolution implementation is not as obvious, so we will outline the motivation and reasoning behind our design choices.

There are several ways to implement convolutions efficiently.
Our goal is to provide performance as close as possible to matrix multiplication, while using no auxiliary memory.
GPU memory is high bandwidth, but low capacity, and is therefore a scarce resource.
When training deep networks, ideally the GPU memory should be filled with data, parameters, and neuron responses, not auxiliary data structures needed by the convolution algorithm.
Several approaches to computing convolutions require large auxiliary data structures, and therefore we do not consider these approaches for cuDNN.

One approach is to lower the convolutions into a matrix multiplication, following \cite{Chellapilla2006}.
This can be done by reshaping the filter tensor $F$ into a matrix $F_m$ with dimensions $K \times CRS$, and gathering a data matrix by duplicating the original input data into a matrix $D_m$ with dimensions $CRS \times NPQ$.
The computation can then be performed with a single matrix multiply to form an output matrix $O_m$ with dimension $K \times NPQ$.
\begin{figure}[h]
\centering
\includegraphics[width=4in]{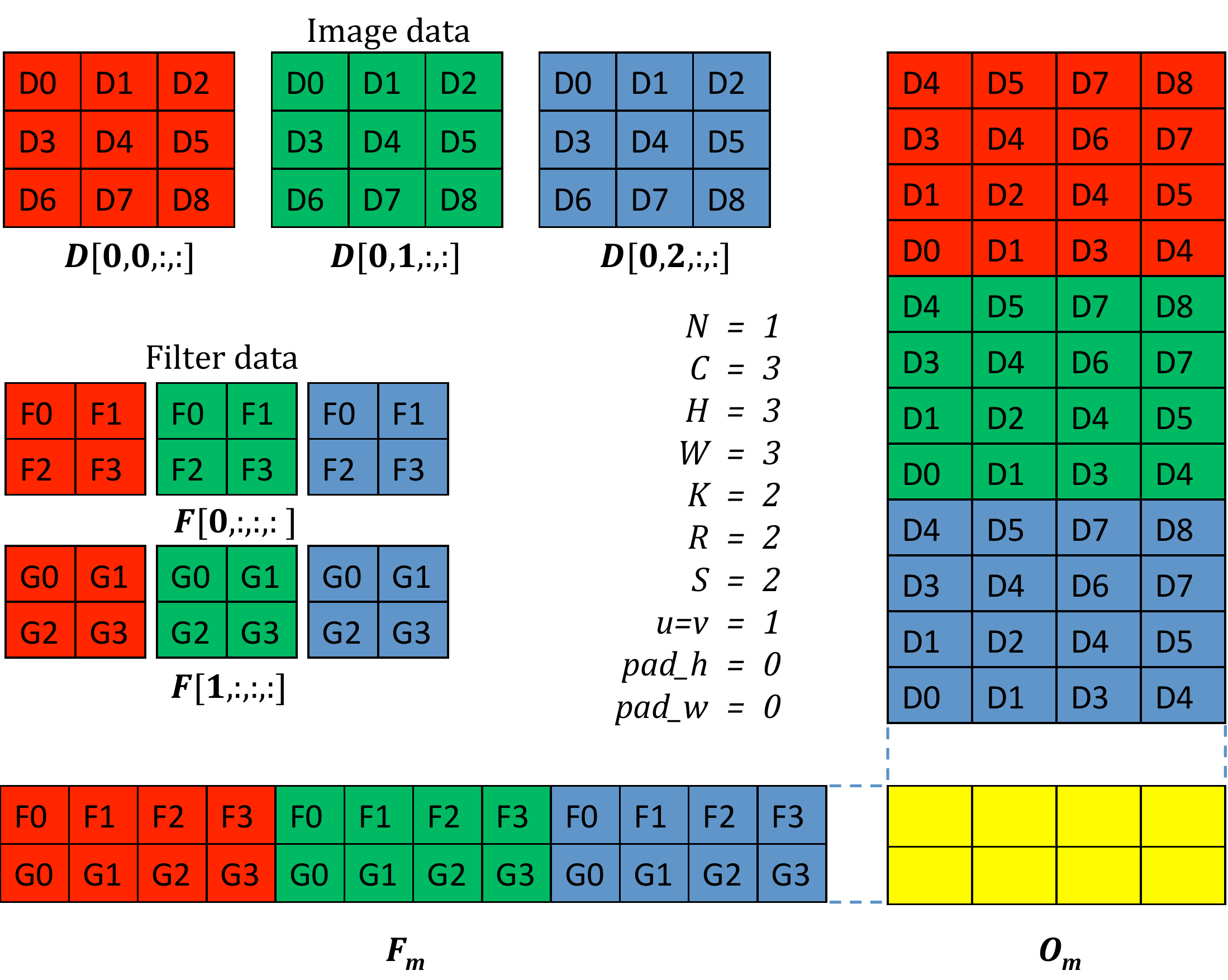}
\caption{Convolution lowering}
\label{fig:lower}
\end{figure}

Figure \ref{fig:lower} illustrates how a simple convolution can be lowered to a matrix multiplication.
The colors in this illustration represent the input feature maps, and elements of $D$ and $F$ are uniquely labeled in the illustration so as to show how each participates in forming $D_m$ and $F_m$.
The filter matrix $F_m$ has dimensions $K \times CRS = 2 \times 12$, while the data matrix $D_m$ has dimensions $CRS \times NPQ = 12 \times 4$.
Note that each element of $D$ is duplicated up to $RS=4$ times in $D_m$.
The output matrix $O_m$ has dimensions $K \times NPQ = 2 \times 4$.

Lowering convolutions to matrix multiplication can be efficient, since matrix multiplication is highly optimized.
Matrix multiplication is fast because it has a high ratio of floating-point operations per byte of data transferred.
This ratio increases as the matrices get larger, meaning that matrix multiplication is less efficient on small matrices.
Accordingly, this approach to convolution is most effective when it creates large matrices for multiplication.
As we mentioned earlier, the sizes of $D_m$ and $F_m$ depend on products of the parameters to the convolution, not the parameters themselves.
This means that performance using this approach can be very consistent, since the algorithm does not care if one of the parameters is small, as long as the {\em product} is large enough.
For example, it is often true that in early layers of a convolutional network, $C$ is small, but $R$ and $S$ are large, while at the end of the network, $C$ is large, but $R$ and $S$ are small.
However, the product $CRS$ is usually fairly large for all layers, so performance can be consistently good.
The disadvantage of this approach is that forming $D_m$ involves duplicating the input data up to $RS$ times, which can require a prohibitively large temporary allocation.
To work around this, implementations sometimes materialize $D_m$ piece by piece, for example, by calling matrix multiplication iteratively for each element of the mini-batch.
However, this limits the parallelism in the implementation, and can lead to cases where the matrix multiplications are too small to effectively utilize the GPU.
This approach also lowers the computational intensity of the convolutions, because $D_m$ must be written and read, in addition to reading $D$ itself, requiring significantly more memory traffic as a more direct approach.
Accordingly, we opt not to use this implementation directly, although as we will explain, our implementation is related.

Another approach is to use the Fast Fourier Transform to compute the convolution.
The FFT can significantly lower the work complexity of the convolutions, and with clever engineering can be used effectively for deep neural networks \cite{Mathieu2013}.
However, the FFT based approach uses a significant amount of temporary memory, since the filters must be padded to be the same size as the inputs.
This is especially costly when the filters are small compared to the images, which often happens in the first few layers of a convolutional network.
Additionally, the FFT based approach does not perform efficiently when the striding parameters $u$ and $v$ are greater than $1$, which is common in many state-of-the art networks, such as the early layers in \cite{Sermanet2013} and \cite{Szegedy2014}.
Striding reduces the computational work of the convolutions by a factor of $uv$, by computing only a sparse subset of the output.
However, the nature of the FFT algorithm is such that computing pruned FFTs is a non-trivial task, and often is slower than computing the dense FFT, followed by an additional subsampling step.
Due to these drawbacks, we opted to forgo the FFT approach, although we agree that in some cases it is useful.

Another common approach is to compute the convolutions directly.
This can be very efficient, but requires a large number of specialized implementations to handle the many corner cases implicit in the 11-dimensional parameter space of the convolutions.
Implementations following this approach are often well-optimized for convolutions in certain parts of the parameter space, but perform poorly for others.
For example, cuda-convnet2 \cite{cudaconvnet2-2014} performs well when batch sizes are large, but poorly once batch size falls to 64 or below.
Optimizing and maintaining all these specializations is a difficult task.
As we envision this library being maintained for some time, and being ported to yet-to-be-conceived future architectures, we searched for something simpler that would perform more robustly across the parameter space and be easier to port to new architectures.

\subsection{Our approach}

NVIDIA provides a matrix multiplication routine that achieves a substantial fraction of floating-point throughput on GPUs.
The algorithm for this routine is similar to the algorithm described in \cite{Tan2011}.
Fixed sized submatrices of the input matrices $A$ and $B$ are successively read into on-chip memory and are then used to compute a submatrix of the output matrix $C$.
We compute on tiles of $A$ and $B$ while fetching the next tiles of $A$ and $B$ from off-chip memory into on-chip caches and other memories.
This technique hides the memory latency associated with the data transfer, allowing the matrix multiplication computation to be limited only by the time it takes to perform the arithmetic.

As we discussed earlier, convolutions can be lowered onto matrix multiplication.
This approach provides simplicity of implementation as well as consistency of performance across the parameter space, although materializing the lowered matrix in memory can be costly.
Our solution follows this approach, but we avoid the problems with materializing the lowered matrix in memory by lazily materializing $D_m$ in on-chip memory only, rather than by materializing it in off-chip memory before calling a matrix multiplication routine.
Since the tiling required for the matrix multiplication routine is independent of any parameters from the convolution, the mapping between the tile boundaries of $D_m$ and the convolution problem is non-trivial.
Accordingly, our approach entails computing this mapping and using it to load the correct elements of $A$ and $B$ into on-chip memories.
This happens dynamically as the computation proceeds, which allows our convolution algorithm to exploit optimized infrastructure for matrix multiplication.
We require additional indexing arithmetic compared to a matrix multiplication, but fully leverage the computational engine of matrix multiplication to perform the work.
After the computation is complete, we perform the required tensor transposition to store the result in the user's desired data layout.

Computing the additional indexing requires repeated calculations of integer division and modulus operations by launch-time constant divisors.
We make use of the algorithm for integer division and modulus presented in \cite{Warren2003} to transform these costly operations into integer multiplies and shifts, and thus reduce the indexing overhead required by our approach.

\subsection{Performance}
The convolution routines in cuDNN provide competitive performance with zero auxiliary memory required.
\begin{figure}
\centering
\includegraphics[width=4in]{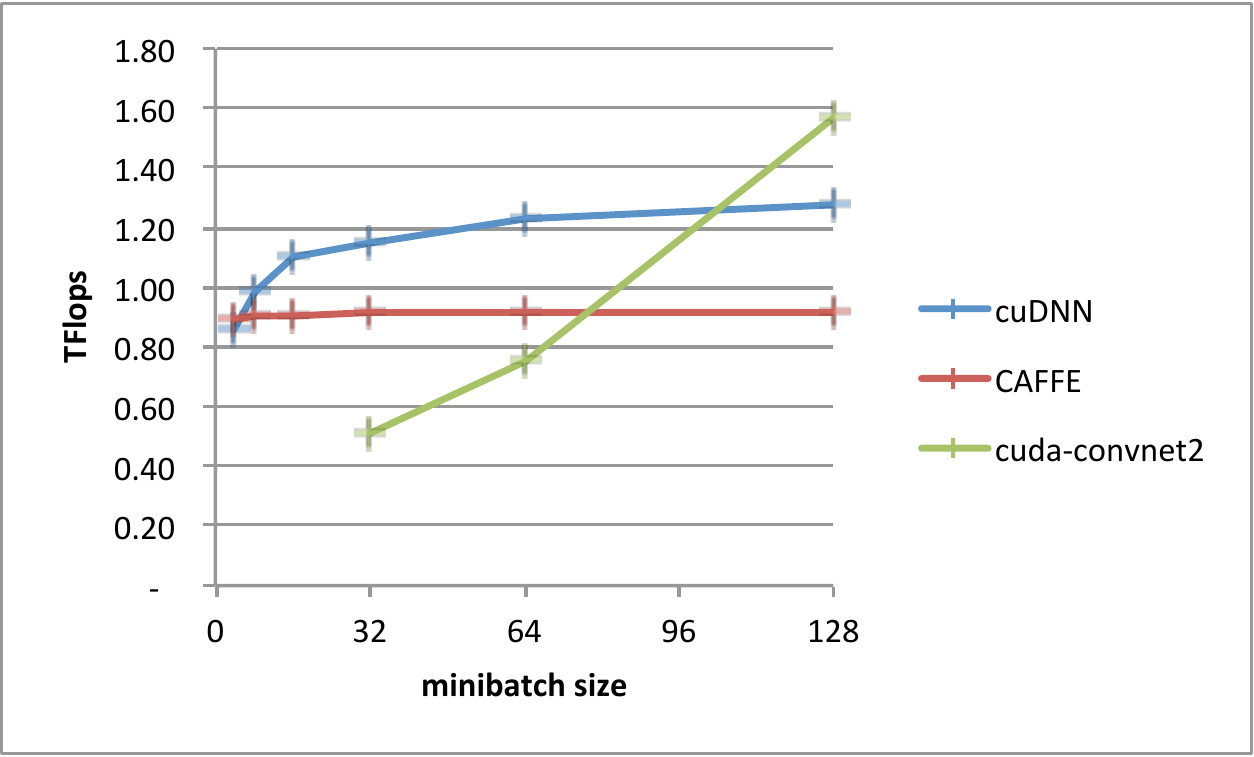}
\caption{Comparative Performance}
\label{fig:comp_perf}
\end{figure}

\begin{table}
\centering
\begin{tabular}{llllllllllll}
 & $N$ & $C$ & $H$ & $W$ & $K$ & $R$ & $S$ & $u$ & $v$ & $pad\_h$ & $pad\_w$ \\ \hline
Layer 1 & 128 & 3 & 128 & 128 & 96 & 11 & 11 & 1 & 1 & 0 & 0 \\
Layer 2 & 128 & 96 & 64 & 64 & 128 & 9 & 9 & 1 & 1 & 0 & 0 \\
Layer 3 & 128 & 128 & 32 & 32 & 128 & 9 & 9 & 1 & 1 & 0 & 0 \\
Layer 4 & 128 & 128 & 16 & 16 & 128 & 7 & 7 & 1 & 1 & 0 & 0 \\
Layer 5 & 128 & 128 & 13 & 13 & 384 & 3 & 3 & 1 & 1 & 0 & 0
\end{tabular}
\caption{Convolutional layer collection}
\label{tab:layers}
\end{table}

Figure \ref{fig:comp_perf} shows the performance on an NVIDIA Tesla K40 of three convolution implementations: cuDNN, Caffe, and cuda-convnet2.
We evaluated these implementations using the layer configurations shown in table \ref{tab:layers}, which are commonly used for benchmarking convolution performance \cite{Soumith2014}, and quote average throughput for all these layers.
cuDNN performance ranges from $0.8\times$ to $2.25\times$ that of cuda-convnet2, with an advantage at smaller batch sizes.
Compared to Caffe, cuDNN performance ranges from $1.0\times$ to $1.41\times$.
Importantly, even with a small mini-batch size of only 16, cuDNN performance is still 86\% of maximum performance, which shows that our implementation performs well across the convolution parameter space.

\begin{table}
\centering
\begin{tabular}{l|ll|ll}
& K40 (TFlops) & GTX980 (TFlops) & K40 (\%) & GTX980 (\%) \\ \hline
Layer 1 & 0.99 & 1.46 & 23\% & 30\% \\
Layer 2 & 1.52 & 2.49 & 35\% & 51\% \\
Layer 3 & 1.51 & 2.42 & 35\% & 49\% \\
Layer 4 & 1.38 & 2.19 & 32\% & 45\% \\
Layer 5 & 1.01 & 1.87 & 24\% & 38\% \\
\end{tabular}
\caption{Performance portability}
\label{tab:perf_port}
\end{table}

Table \ref{tab:perf_port} illustrates cuDNN's performance portability across GPU architectures.
The Tesla K40 is built using the Kepler architecture, and has a peak single-precision throughput of 4.29 TFlops.
The Geforce GTX 980 is built using the newer Maxwell architecture, and has a peak single-precision throughput of 4.95 TFlops.
cuDNN performance ranges from 23-35\% of peak on the Tesla K40, and from 30-51\% of peak on the GTX 980.
This data illustrates how cuDNN provides performance portability across GPU architectures, with no need for users to retune their code as GPU architectures evolve.

\section{Caffe Integration}

Caffe is a deep learning framework developed with expression, speed, and modularity in mind.
Its architecture mirrors the natural modularity of deep networks as compositional models made from a collection of inter-connected layers.
A deep network is defined layer-by-layer in a plaintext schema.
Each layer type is implemented according to a simple protocol of setup, forward, and backward steps to encapsulate engineering details.
Data and derivatives flow through layers and between the host and device according to the framework's unified memory interface that handles allocation and communication.
cuDNN integration raises the speed and memory efficiency of the framework without sacrificing expression or modularity.

\subsection{Development}

Integration is made simple by the self-contained design of the cuDNN handles, descriptors, and function calls together with the modularity of the framework.
The core Caffe framework is unaltered, preserving the network, layer, and memory interfaces.
Changes were isolated to new layer definitions and implementations, helper functions for descriptors, and the corresponding tests.
The patch is almost purely additive.

In Caffe each type of model operation is encapsulated in a layer.
Layer development comprises declaring and implementing a layer class, defining the layer in the protocol buffer model schema, extending the layer factory, and including tests.
Computations are carried out through the layer protocol of setup, forward, and backward steps.
cuDNN layers fit this same scheme.
Library handles and descriptors are configured in setup while forward and backward calls are made in the respective layer methods.
The cuDNN primitives yield concise layer implementations.
The cuDNN layers are drop-in replacements to their standard Caffe counterparts.

Caffe has a standard array and memory interface, called a blob, for storing and communicating data on the host and device.
Blobs hold data, gradients, and parameters.
In particular, layer inputs and outputs are held in $N \times C \times H \times W$ dimensional blobs.
cuDNN tensor and filter descriptors are trivially constructed from blob arrays through its flexible support for dimension and stride.
By coordinating memory solely at the descriptor, Caffe retains control over memory and communication for efficiency.

The Caffe$+$cuDNN convolution layer exploits the reduced memory consumption of cuDNN to speed up execution.
The forward pass parallelizes the computation of group convolution (filtering with sets of filters with restricted channel connectivity).
The backward pass parallelizes the computation of the gradients with respect to the bias, filter weights, and bottom data.

\begin{table}
\centering
\begin{tabular}{llll}
& Caffe (seconds) & Caffe$+$cuDNN (seconds) & Speedup \\ \hline
Backward Propagation & 156270 & 120651 & 1.30 \\
Forward Propagation & 109310 & 75330 & 1.45 \\
Testing & 21600 & 14529 & 1.49 \\
Update & 2926 & 2920 & 1.00 \\
Overall & 290106 & 213431 & 1.36
\end{tabular}
\caption{Caffe Performance}
\label{tab:caffe}
\end{table}

Table \ref{tab:caffe} illustrates the performance improvements gained from integrating cuDNN into Caffe.
Overall, training time for 200 iterations improved by 36\%, when training the bvlc\_reference\_caffenet model, using cuDNN R1 on an NVIDIA Tesla K40.

\subsection{User Experience}

cuDNN computation is transparent to the user through drop-in integration.
The model schema and framework interfaces are completely unchanged.
Setting a single compilation flag during installation equips Caffe with cuDNN layer implementations and sets cuDNN as the default computation engine.
Layers automatically fall back to the standard Caffe functionality in all cases outside of the current scope of cuDNN.
Model execution can be tuned through per-layer \textsc{engine} parameters that select the implementation.
The workflow for the user is virtually identical.

Future cuDNN releases will be integrated in like fashion.

\section{Baidu Integration}
Several deep learning projects at Baidu have integrated cuDNN.
For example, it has been integrated into PADDLE, Baidu's internal deep learning framework.
On a Tesla K10 GPU, we observed that performance on a benchmark set of convolutional layers improves by 30\% on average over an implementation that lowers convolution to matrix multiply.

We are also using cuDNN in other domains besides image processing, such as speech and language.
cuDNN's ability to convolve non-square inputs with asymmetric padding is particularly useful for these other domains.
In our experience, cuDNN has reduced memory consumption compared with matrix multiplication, which has enabled us to use larger models and larger mini batches.
cuDNN was simple to integrate into our code because we didn't need to change our data structure layout, thanks to cuDNN's flexible interface.

\section{Future Work}
We are considering several avenues for expanding the performance and functionality of cuDNN.
Firstly, although our convolution routines are competitive with other available implementations, more work remains to bring performance up to that attained by matrix multiplication.
Over time, we hope to shrink this gap.
Secondly, we envision adding support for other primitives; for example: 1D and 3D convolutions would be useful for speech, language processing, and video applications, among others.
Local Receptive Field computations, which are similar to convolutions, but with untied weights, are also useful and could be added to cuDNN.
Finally, we would like this library to help people use multiple GPUs to accelerate training.

\section{Conclusion}
This paper presents cuDNN, a library for deep learning primitives.
We presented a novel implementation of convolutions that provides reliable performance across a wide range of input sizes, and takes advantage of highly-optimized matrix multiplication routines to provide high performance, without requiring any auxiliary memory.
We also provide a set of routines that allow users to train and evaluate complete deep neural networks without the need to write parallel code manually.
As parallel architectures continue to evolve, such libraries will provide increasing value to the machine learning community.
The library is available at \cite{cuDNN2014}, and we welcome feedback at \url{cuDNN@nvidia.com}.

\bibliographystyle{plain}
\bibliography{paper}

\begin{thebibliography}{10}

\bibitem{Soumith2014}
convnet-benchmarks.
\newblock \url{https://github.com/soumith/convnet-benchmarks}, 2014.

\bibitem{NetlibBLAS}
Netlib {BLAS}.
\newblock \url{http://www.netlib.org/blas/}, 2014.

\bibitem{cuDNN2014}
{NVIDIA} {cuDNN} - {GPU} accelerated deep learning.
\newblock \url{https://developer.nvidia.com/cuDNN}, 2014.

\bibitem{Bengio2012}
Yoshua Bengio, Aaron~C. Courville, and Pascal Vincent.
\newblock Unsupervised feature learning and deep learning: {A} review and new
  perspectives.
\newblock {\em CoRR}, abs/1206.5538, 2012.

\bibitem{Bergstra2010}
James Bergstra, Olivier Breuleux, Fr{\'e}d{\'e}ric Bastien, Pascal Lamblin,
  Razvan Pascanu, Guillaume Desjardins, Joseph Turian, David Warde-Farley, and
  Yoshua Bengio.
\newblock Theano: a cpu and gpu math expression compiler.
\newblock In {\em SciPy}, volume~4, page~3, 2010.

\bibitem{Chellapilla2006}
Kumar Chellapilla, Sidd Puri, Patrice Simard, et~al.
\newblock High performance convolutional neural networks for document
  processing.
\newblock In {\em Workshop on Frontiers in Handwriting Recognition}, 2006.

\bibitem{Coates2013}
Adam Coates, Brody Huval, Tao Wang, David Wu, Bryan Catanzaro, and Andrew Ng.
\newblock Deep learning with {COTS} {HPC} systems.
\newblock In {\em ICML}, pages 1337--1345, 2013.

\bibitem{Collobert2011}
Ronan Collobert, Koray Kavukcuoglu, and Cl{\'e}ment Farabet.
\newblock Torch7: A matlab-like environment for machine learning.
\newblock In {\em BigLearn, NIPS Workshop}, 2011.

\bibitem{Dahl2012}
George~E Dahl, Dong Yu, Li~Deng, and Alex Acero.
\newblock Context-dependent pre-trained deep neural networks for
  large-vocabulary speech recognition.
\newblock {\em Audio, Speech, and Language Processing, IEEE Transactions on},
  20(1):30--42, 2012.

\bibitem{Hinton2012}
Geoffrey Hinton, Li~Deng, Dong Yu, George~E Dahl, Abdel-rahman Mohamed, Navdeep
  Jaitly, Andrew Senior, Vincent Vanhoucke, Patrick Nguyen, Tara~N Sainath,
  et~al.
\newblock Deep neural networks for acoustic modeling in speech recognition: The
  shared views of four research groups.
\newblock {\em Signal Processing Magazine, IEEE}, 29(6):82--97, 2012.

\bibitem{Jia2014}
Yangqing Jia, Evan Shelhamer, Jeff Donahue, Sergey Karayev, Jonathan Long, Ross
  Girshick, Sergio Guadarrama, and Trevor Darrell.
\newblock Caffe: Convolutional architecture for fast feature embedding.
\newblock {\em arXiv preprint arXiv:1408.5093}, 2014.

\bibitem{cudaconvnet2-2014}
Alex Krizhevsky.
\newblock cudaconvnet2.
\newblock \url{https://code.google.com/p/cuda-convnet2/}, 2014.

\bibitem{Krizhevsky2012}
Alex Krizhevsky, Ilya Sutskever, and Geoffrey~E Hinton.
\newblock Imagenet classification with deep convolutional neural networks.
\newblock In {\em NIPS}, pages 1097--1105, 2012.

\bibitem{Lecun1998}
Yann LeCun, L{\'e}on Bottou, Yoshua Bengio, and Patrick Haffner.
\newblock Gradient-based learning applied to document recognition.
\newblock {\em Proceedings of the IEEE}, 86(11):2278--2324, 1998.

\bibitem{Maas2014}
Andrew~L Maas, Awni~Y Hannun, Daniel Jurafsky, and Andrew~Y Ng.
\newblock First-pass large vocabulary continuous speech recognition using
  bi-directional recurrent dnns.
\newblock {\em arXiv preprint arXiv:1408.2873}, 2014.

\bibitem{Mathieu2013}
Michael Mathieu, Mikael Henaff, and Yann LeCun.
\newblock Fast training of convolutional networks through ffts.
\newblock {\em arXiv preprint arXiv:1312.5851}, 2013.

\bibitem{ILSVRC2014}
Olga Russakovsky, Jia Deng, Hao Su, Jonathan Krause, Sanjeev Satheesh, Sean Ma,
  Zhiheng Huang, Andrej Karpathy, Aditya Khosla, Michael Bernstein,
  Alexander~C. Berg, and Li~Fei-Fei.
\newblock Imagenet large scale visual recognition challenge, 2014.

\bibitem{Sermanet2013}
Pierre Sermanet, David Eigen, Xiang Zhang, Micha{\"e}l Mathieu, Rob Fergus, and
  Yann LeCun.
\newblock Overfeat: Integrated recognition, localization and detection using
  convolutional networks.
\newblock {\em arXiv preprint arXiv:1312.6229}, 2013.

\bibitem{Szegedy2014}
Christian Szegedy, Wei Liu, Yangqing Jia, Pierre Sermanet, Scott Reed, Dragomir
  Anguelov, Dumitru Erhan, Vincent Vanhoucke, and Andrew Rabinovich.
\newblock Going deeper with convolutions.
\newblock {\em arXiv preprint arXiv:1409.4842}, 2014.

\bibitem{Tan2011}
Guangming Tan, Linchuan Li, Sean Treichler, Everett Phillips, Yungang Bao, and
  Ninghui Sun.
\newblock Fast implementation of {DGEMM} on {Fermi} {GPU}.
\newblock In {\em Supercomputing 2011}, SC '11, pages 35:1--35:11, New York,
  NY, USA, 2011. ACM.

\bibitem{Warren2003}
Henry~S. Warren.
\newblock {\em Hacker's Delight}.
\newblock Addison-Wesley Professional, 2002.

\end{thebibliography}

\end{document}